\documentclass[a4paper,conference]{IEEEtran}
\usepackage{breqn}
\usepackage{amsfonts}
\usepackage{soul}
\usepackage{graphicx}
\usepackage{amsmath}
\usepackage{microtype}
\usepackage{hyperref} 
\usepackage{booktabs}
\usepackage{multirow}
\usepackage{pifont}
\usepackage{float}
\usepackage{xcolor}
\usepackage{bm}
\usepackage{multicol}
\usepackage{array}
\usepackage{comment}
\usepackage{hyperref}
\usepackage{mwe} % to get dummy images
\usepackage{enumitem}
\usepackage[normalem]{ulem}

\newcommand{\cmark}{\ding{51}}%
\newcommand{\xmark}{\ding{55}}%

\ifCLASSINFOpdf
\else
\fi
\begin{document}
\title{3DConvCaps: 3DUnet with Convolutional Capsule Encoder for Medical Image Segmentation}


%%%%% Start Authors
\author{\IEEEauthorblockN{Minh Tran, Viet-Khoa Vo-Ho, Ngan T.H. Le}
\IEEEauthorblockA{\large{Dept. of CSCE, 
University of Arkansas,
Fayetteville, USA}\\
% \normalise
\texttt{\small\{minht, khoavoho, thile\}@uark.edu}}
% \and
% \IEEEauthorblockN{Viet-Khoa Vo-Ho}
% \IEEEauthorblockA{Dept. of CSCE\\
% University of Arkansas\\
% Fayetteville, USA\\
% Email: khoavoho@uark.edu}
% \and
% \IEEEauthorblockN{Ngan T.H. Le}
% \IEEEauthorblockA{Dept. of CSCE\\
% University of Arkansas\\
% Fayetteville, USA\\
% Email: thile@uark.edu}
}


\maketitle

\begin{abstract}
Convolutional Neural Networks (CNNs) have achieved promising results in medical image segmentation. However, CNNs require lots of training data and are incapable of handling pose and deformation of objects. Furthermore, their pooling layers tend to discard important information such as positions as well as CNNs are sensitive to rotation and affine transformation. Capsule network is a recent new architecture that has achieved better robustness in part-whole representation learning by replacing pooling layers with dynamic routing and convolutional strides, which has shown potential results on popular tasks such as digit classification and object segmentation. In this paper, we propose a 3D encoder-decoder network with Convolutional Capsule Encoder (called 3DConvCaps) to learn lower-level features (short-range attention) with convolutional layers while modeling the higher-level features (long-range dependence) with capsule layers. Our experiments on multiple datasets including iSeg-2017, Hippocampus, and Cardiac demonstrate that our 3D 3DConvCaps network considerably outperforms previous capsule networks and 3D-UNets. We further conduct ablation studies of network efficiency and segmentation performance under various configurations of convolution layers and capsule layers at both contracting and expanding paths. The implementation is available at: \hyperlink{https://github.com/UARK-AICV/3DConvCaps}{https://github.com/UARK-AICV/3DConvCaps}

\end{abstract}


% \fi
\IEEEpeerreviewmaketitle

\section{Introduction}
\label{sec:intro}
Medical image segmentation (MIS) aims to partition image pixels to identify anatomical structures of interest. MIS has been applied in numerous clinical applications in healthcare varying from computer-aided diagnosis, to treatment planning to survival analysis. With recent advances in machine learning, i.e. convolutional neural networks (CNNs), MIS has become increasingly promising.
Since the introduction of UNet~\cite{ronneberger2015u,cciccek20163d}, UNet-based CNNs have achieved impressive performance in various modalities of MIS, e.g. brain tumor~\cite{brain_le_2018, le2021multi, ho2021point}, infant brain \cite{hoang2021dam, le2021offset}, liver tumor~\cite{bilic2019liver}, optic disc~\cite{ramani2020improved}, retina \cite{le2021narrow},  lung~\cite{nguyen20213ducaps}, and cell~\cite{moshkov2020test}, etc. However, each feature map in such encoder-decoder architecture only contains information about the presence of the feature, and the network relies on fixed learned weight matrix to link features between layers. Thus, such models cannot generalize well to unseen changes in the input image and usually perform poorly in that case. Furthermore, pooling layer in CNNs, which summarizes features in a local window, discards important spatial information such as pose and part-whole relationship. Furthermore, CNNs are sensitive to rotation and affine transformation \cite{nguyen20213ducaps}. 
\begin{table}[t]
\centering
\caption{Network comparison between our 3DConvCaps with existing Capsule-based MIS methods. Conv. and Caps present Convolutional blocks and Capsule blocks, respectively.}
\begin{tabular}{lccc}
\toprule
 & Input & Encoder & Decoder \\ \hline
2D SegCaps\cite{lalonde2018capsules} &  2D   & Caps.     &    Caps.    \\
{Multi-SegCaps \mbox{\cite{survarachakan2020capsule}}} &  2D   & Caps.     &    Caps.    \\
3D-SegCaps\cite{nguyen20213ducaps} &  3D  &   Caps.  &     Caps.   \\ 
3D-UCaps\cite{nguyen20213ducaps} &  3D  &  Caps.   &   Conv.      \\ 
SS-3DUCaps\cite{tran2022ss} &  3D  &   Caps.  &   Conv.     \\ \hline
\textbf{Our 3DConvCaps}  &  3D  & Conv. \& Caps.     &    Conv.     \\ 
\bottomrule
\end{tabular}
\label{tb:compare}
\end{table}

To overcome the aforementioned limitations by CNNs, Sabour et al.~\cite{sabour2017dynamic} proposed Capsule network, which encodes the part-whole relationships including locations, scale locations, orientations between various parts of objects, to achieve viewpoint equivariance. To obtain such objectives, Capsule models the relationship between part features of the objects by a novel optimization mechanism, i.e., dynamic routing. In such mechanism, contributions of parts to a whole object are weighted differently at both training and testing. Capsule was originally applied to {natural} image recognition, it is recently extended to medical imaging, {where it helps with the problem of small amounts of annotated data and class-imbalance \mbox{\cite{jimenez2018capsule}}.
Moreover, capsuled networks are also applied
on medical image segmentation task \mbox{\cite{lalonde2018capsules,survarachakan2020capsule,lalonde2021capsules,nguyen20213ducaps}} with promising results}. Capsule-based MIS was first introduced by 2D SegCaps \cite{lalonde2018capsules}, which suggested that Capsule-based approach in MIS {has
better performance, by allowing for preservation of part-whole relationships in the data}. However, these networks are mainly designed for 2D still images, and they performs poorly when being applied to volumetric data because of missing temporal information. Later, Nguyen, et al. \cite{nguyen20213ducaps} proposed 3D-UCaps, in which the encoder is built on 3D Capsule blocks and decoder is built on 3D CNNs. The empirical results have shown that 3D-UCaps provide better performance is more robust than 2D SegCaps. However, they both are highly dependent on some random phenomena such as sampling order or weight initialization. Recently, Tran et al. \cite{tran2022ss} adopted self-supervised learning to 3D-UCaps and proposed SS-3DCapsNet. In such 3D Capsule-based MIS approaches \cite{nguyen20213ducaps, tran2022ss}, Capsule blocks are applied at every layers in the network encoder i.e. contracting path. Even Capsule-based MIS methods contain less number of parameters, their inferences are still time consuming because of optimizing procedure of dynamic routing. Furthermore, it has been proven that convolutional layers are able to model short-range attention with local information around the receptive fields whereas capsule layers are capable of encoding the part-whole relationships, which present long-range dependence. 

{Based on the above progress, our motivation comes from the success of 3D-UCaps \mbox{\cite{nguyen20213ducaps}}
% (oral, MICCAI 2021)
. It has been proved by 3D-UCaps \mbox{\cite{nguyen20213ducaps}} that 3D-UCaps is not only a SOTA capsule-based medical image segmentation approach but also achieves competitive performance with plausible model size compared with CNN-based approaches. However, like other capsule-based approaches, inference time is a shortcoming in 3D-UCaps because of the dynamic routing procedure. Our ablation study shows that capsule blocks work effectively at high-level feature to model global relationship, whereas convolutional block is able to proficiently learn local feature. Therefore, we improve 3D-UCaps by redesigning the contracting path, which starts with convolutional blocks to capture low-level features and then capsule blocks to handle high-level features. The objectives of our proposed 3DConvCaps are two-fold: i) aims to extract better visual representation while modeling the part-whole relations; ii) reduce inference time while keep being a plausible model size.} The comparison between our 3DConvCaps and the existing Capsule-based MIS methods is given in Table \ref{tb:compare}.

\section{Related Work}

\subsection{CNN-based MIS}
CNN-based approaches with encoder-decoder architectures have shown excellent performance in MIS. Besides 2.5D methods \cite{liu20183d, le2018deep, xia20203d}, in which tri-planar architectures are sometimes used to combine three-view slices for each voxel, 3D approaches \cite{bakas2018identifying, heller2019kits19}, {\mbox{\cite{zheng2021hierarchical}}} have achieved state-of-the-art results in various medical semantic segmentation tasks. 
Early methods include the standard Unet~\cite{ronneberger2015u}, Vnet~\cite{VNet}, and then DeepMedic~\cite{deepmedic}, which improves robustness with multi-scale segmentation. Following this, many subsequent works follow this encoder-decoder structure, experimenting with dense connections, skip connections, residual blocks, and other types of architectural additions to improve segmentation accuracies for particular medical imaging applications. Although encoder-decoder architectures have powerful representation learning capabilities, their performance in learning long-range dependencies and model weak boundary objects as well as imbalanced class distribution is limited \cite{hatamizadeh2022unetr}. Some efforts have used atrous convolutional layers \cite{brain_Le_2021_2, gu2019net}, active contour \cite{le2021narrow, le2021offset}, ensembled model \cite{Ensemble} to address the above problems. 

\subsection{Capsule-based MIS}
Similar to CNN-based MIS, Capsule-based MIS \cite{jimenez2018capsule, lalonde2018capsules, lalonde2021capsules, nguyen20213ducaps, tran2022ss} 
are a hierarchical encoder-decoder architecture. However, Capsule-based MIS approaches build tensors by grouping multiple feature channels instead of applying a non-linearity to the scalar outputs of filters in CNNs. In general, Capsule-based networks are formed by two layers: a first primary layer, capturing low-level features, followed by a specialized secondary layer, predicting both the presence and pose of an object in the image. Furthermore, each layer in Capsule aims to learn a set of entities (i.e., parts or objects) and represent them in a high-dimensional form \cite{sabour2017dynamic}. An important assumption in Capsule is the entity in the previous layer is a simple feature and based on an agreement in their votes, the complex feature in the next layer will be activated or not. 

SegCaps~\cite{lalonde2018capsules, jimenez2018capsule} are the first Capsule-based MIS methods, which are designed as a 2D encoder-decoder architecture. Such networks have made a great improvement to expand the use of CapsNet in image classification to the task of object segmentation. Especially, they demonstrate the increased generalization ability of Capsule-based MIS  when dealing with the limited amount of data and class-imbalance. However, 2D SegCaps~\cite{lalonde2018capsules, jimenez2018capsule} ignore temporal information, and thus, they perform poorly when being applied to volumetric data. To address such limitations, \cite{nguyen20213ducaps, tran2022ss} propose  3D-UCaps and SS-3DCapsNet for volumetric-wise MIS by taking the advantages of both Capsule and CNN into consideration. In 3D-UCaps~\cite{nguyen20213ducaps} the encoder is designed by 3D Capsule blocks whereas the decoder is designed by convolutional blocks. Like other deep learning networks, 3D-UCaps is dependent on initial weights. Recently, Tran et al., propose  SS-3DCapsNet~\cite{tran2022ss} which is an extension of 3D-UCaps where the initial weights is learnt by a self-supervised learning mechanism. Even thought both 3D-UCaps~\cite{nguyen20213ducaps} and SS-3DCapsNet~\cite{tran2022ss} show good performance with a fewer number layers and parameters on various medical datasets, their inference is still time consumption because of dynamic routing. In this paper, we first investigate the advantage of both 3D CNN-based and Capsule-based MIS. We then propose 3DConvCaps network in which the lower-level feature is attended by 3D convolutional blocks and higher-level feature is modeled by 3D Capsules blocks. The difference between our 3DConvCaps and other existing Capsule-based MIS is shown in Table \ref{tb:compare}.

\begin{figure*}[t]
    \centering
    \includegraphics[width=0.95\textwidth]{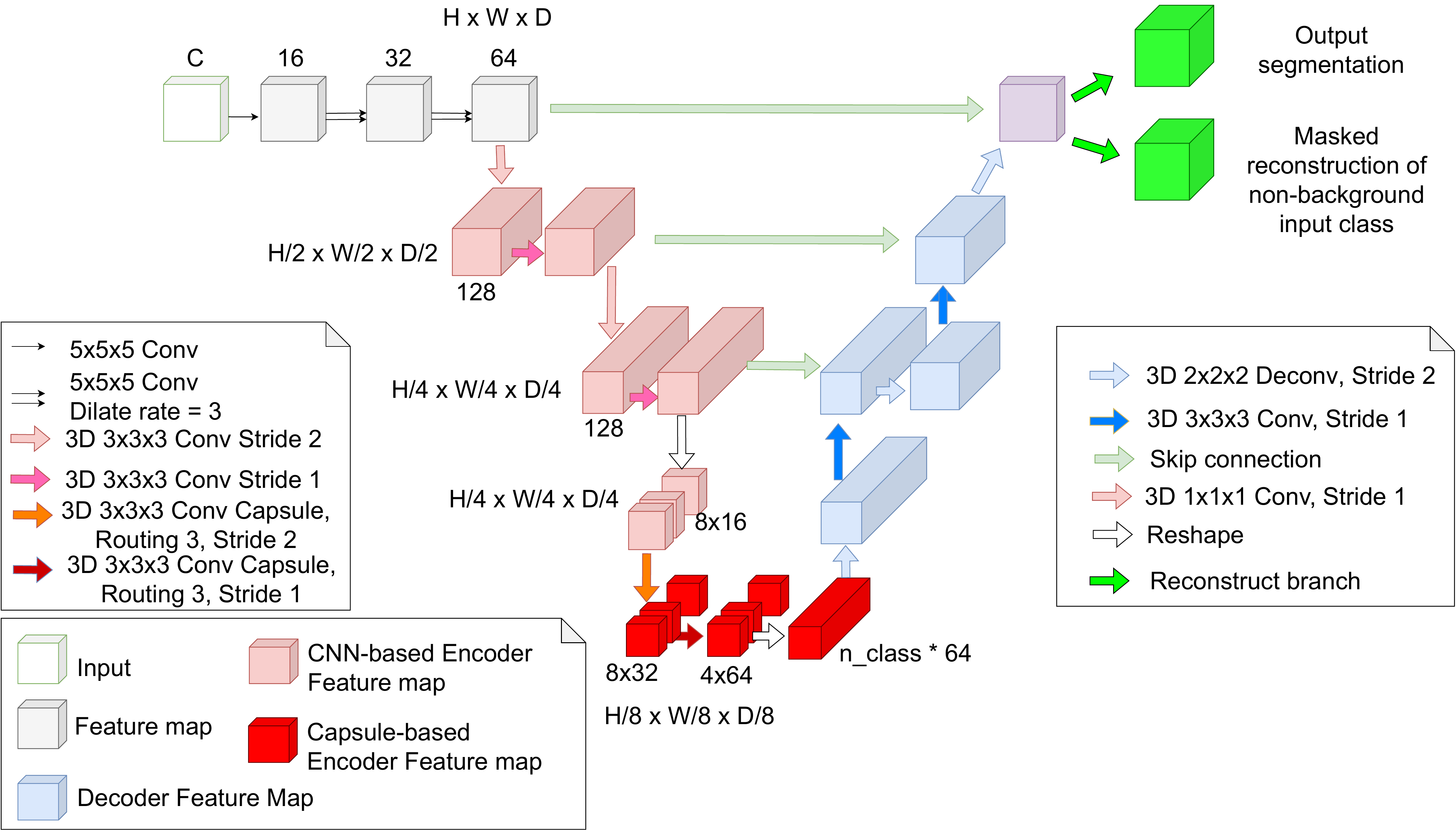}
    \label{fig:network}
    \caption{Overall our proposed 3DConvCaps architecture with three components: Visual feature extraction; ConvCaps encoder {(Conv Encoder and Capsule Encoder)}, and Conv decoder. The number on the blocks indicates the number of channels in the convolution layer and the dimension of capsules in capsule layers.} 
\end{figure*}

\section{Our proposed 3DConvCaps}
\subsection{Capsule Network: Revise}
Capsule networks~\cite{hinton2011transforming} is a new network architecture concept that strengthens feature learning by retaining more information at the aggregation layer for pose reasoning and learning the part-whole relationship, which makes it a potential solution for semantic segmentation and object detection tasks. In Capsule network, a capsule aims to represent an entity: capsule norm indicates the probability that entity is present and capsule direction indicates the configuration that entity is in. Capsule network is recently made practical \cite{sabour2017dynamic} in a CNN that incorporates two layers of capsules with dynamic routing. Capsule networks are subsequently improved with better routing algorithms~\cite{hinton2018matrix, ribeiro2020capsule, gu2020improving}.
An important assumption in Capsule network is that the {entities} in the previous layer are simple objects whereas the {entities} in the next layer are complex objects. The complex objects are formed by simple objects using voting mechanism, i.e dynamic routing. Dynamic routing is then improved by an optimizer \cite{wang2018optimization}, group capsule networks \cite{lenssen2018group}, Expectation-Maximization \cite{hinton2018matrix},  Singular Value Decomposition \cite{bahadori2018spectral}, Variational bayes \cite{ribeiro2020capsule}, Routing-by-agreement mechanism \cite{venkatraman2020learning}, sharing transformation matrices between capsules \cite{gu2020improving}.

\subsection{Network architecture}
Leveraging the intuition by SegCaps \cite{lalonde2018capsules} and 3D-UCaps \cite{nguyen20213ducaps}, we propose 3DConvCaps to tackle the limitation of the existing Capsule-based MIS, that is {time-consuming inference procedure, even when the} model size is small {(as in Table \mbox{\ref{tb:ablation}})}. Our 3DConvCaps is designed as an encoder-decoder architecture while inheriting the merits of both convolutional blocks and capsule blocks. In 3DConvCaps, the lower-level features are encoded by convolutional blocks to capture short-range information whereas the higher-level features are encoded by capsule blocks to learn long-range dependence of part-whole relationship. Similar to 3D-UCaps, our network decoder is design with convolutional blocks. The overall 3DConvCaps network architecture contains three components i.e., (a) Visual feature extraction, (b) ConvCaps feature encoder, and (c) Conv decoder and it is illustrated as in Fig.\ref{fig:network}.

% while solving the problem of model size as well as the rationality of using capsule layers, we designed 3DConvCaps for medical segmentation tasks. The model consists of the three described hereafter stages: (a) Feature extraction, (b) Mix Feature Encoder, and (c) Feature Decoder.\\\\
\noindent   
\textbf{(a) Visual Feature Extraction:} In this component, the visual features is extracted from the original volume image size $H\times W\times D \times C$ via three dilated convolutional layers with kernel size $5 \times 5 \times 5$ and dilate rate set to 1, 3, and 3. The number of channels is increased from 16 to 32 then 64 and the output from this component is a feature map of size $H \times W \times D \times 64$.

%the number of channels increased from 16 to 32 then 64, kernel size $5 \times 5 \times 5$ and dilate rate set to 1, 3, and 3, respectively. 
%Those layers have kernel size of $5\times5\times5$ and the dilated rate of $1, 3$, and $3$ in turn. This extracted visual feature keep the spatial shape while the feature size is richer ($H\times W\times D \times 64$).

\noindent
\textbf{(b) ConvCaps Feature Encoder:} 
This component takes the visual feature size $H \times W \times D \times 64$ from the previous component as its input. The component extracts from lower-level features to higher-level features. Due to the fact the lower level features capture short-range information, convolutional blocks are used. The convolutional layers are defined with the number of channels increased from 64 to 128 and kernel size $3 \times 3 \times 3$. The downsampling is conducted by convolutional layers with the same kernels of $3 \times 3 \times 3$  and strike of 2. The higher-level features capture long-range information, thus, 3D Capsules blocks are used. The number of capsule types in the encoder path of our network is set to $(8, 8)$. To supervisedly learn the model by a margin loss~\cite{sabour2017dynamic}, the number of capsule types in the last capsule layer is equal to the number of categories in the segmentation. In our model, the last capsule layer has the shape of {$H/8 \times W/8 \times D/8 \times C \times A$}, where $C$ is the number of capsule types and $A$ is the dimension of each capsule. 

Unlike SegCaps ~\cite{lalonde2018capsules}, 3D-UCap\cite{nguyen20213ducaps}, SS-3DCaps \cite{tran2022ss}, where capsule layers are utilized at every layer in the contracting path, our 3D ConvCaps adopts convolutional layers at lower level features and capsule layers at higher-level features.

\noindent
\textbf{(c) Conv Decoder: } {It has been proved by 3D-UCaps \mbox{\cite{nguyen20213ducaps}} that using capsules in the expanding path has negligible effects compared to the traditional design while incurring high computational cost due to routing between capsule layers. Thus, there are only convolutional layers used in the expanding path in our network.} This component takes ConvCaps feature from the previous component into consideration and generates the segmentation mask. We first reshape the features to $H \times W \times D \times (C \star A)$. We then pass them through the up-sampling layer, the skip connection and the convolutional layer.

The whole network architecture is described in Fig \ref{fig:network}.

\begin{table*}[th]
    \centering
    \caption{Performance comparison on iSeg-2017 dataset. Train on 10 subjects and test on 13 Subjects. The best is shown in bold and the second best is shown in \underline{underline}. }
    \label{table:iseg}
    \begin{tabular}{ll|l|llll|llll}
    \toprule
        & \multicolumn{1}{c}{\multirow{2}{*}{Method}} & Year & \multicolumn{4}{c}{DSC $\uparrow$}  & \multicolumn{4}{c}{ASD (mm)$\downarrow$}                                        \\ \cmidrule(l){4-11} 
       & &  & WM & GM & CSF & Average &  WM & GM & CSF & Average \\ \cmidrule(l){4-11}
       \multirow{6}{*}{\rotatebox{90}{\shortstack{3D CNNs-based MIS}}} 
    & HyberDense  \cite{dolz2018hyperdense} & 2018& 90.1 & 92.0 & 95.6 & 92.57 & 0.38 & 0.32 & \underline{0.12} & 0.27 \\
    & FC-DenseNet  \cite{hashemi2019exclusive} & 2019& 90.7 & \underline{92.6} & {96.0} & \underline{93.1} & 0.36 & {0.31} & \textbf{0.11} & \underline{0.26} \\
    & D-SkipDenseSeg  \cite{bui2019skip} & 2019& 90.3 & 92.2 & 95.7 & 92.73 & 0.38 & 0.32 & {0.12} & 0.27\\
    & H-DenseNet  \cite{qamar2019multi} & 2019& 90.0 & 92.0 & {96.0} & 92.67 & 0.36 & {0.31} & \textbf{{0.11}} & \underline{0.26} \\
    & FC-Semi-DenseNet1 \cite{SemiDenseNet}& 2020 & 90.0 & 92.0  & {96.0} & 92.67 & 0.38 & 0.35 & 0.14 & 0.29 \\
    & FC-Semi-DenseNet2  \cite{SemiDenseNet}& 2020 & 90.0 & 92.0  & {96.0} & 92.67 & 0.41 & 0.34 & \underline{0.12} & 0.29 \\
    & V-3D-UNet  \cite{qamar2020variant} & 2020& 91.00 & 92.00 & \underline{96.00} & {93.00} & 0.37 & {0.31} & 0.13 & 0.27 \\
    & Non-local U-Net \cite{wang2020non} & 2020 & 91.03 & {92.45} & 95.30 & 92.29 & 0.40 & 0.37  & 0.14 & 0.30\\
    & HyperFusionNet \cite{duan2021multi} & 2021& {90.20} & 87.80 & 93.60 & 90.53 & -- & -- & -- \\
    & APRNet\cite{zhuang2021aprnet} & 2021 & \underline{91.10} & 92.40 & 95.50 & {93.00}& \underline{0.35} & 0.32 & \underline{0.12} & \underline{0.26} \\
    %& DAM-AL \cite{hoang2021dam}& 2022& {92.60} & {93.49} & {95.68} & {93.92} & {0.28} & {0.25} & {0.11} & {0.21} \\
\midrule
\multirow{4}{*}{\rotatebox{90}{\shortstack{Capsule\\-based MIS}}} 
& 2D SegCaps \cite{lalonde2018capsules}  & 2018      & 70.11& 72.66 & 86.31&76.36 & 0.68&0.63 &0.33 &0.55 \\
& 3D-SegCaps \cite{nguyen20213ducaps}     & 2021   & 71.01&73.60 &90.68 &78.43
 &0.52 &0.49 &0.23 &  0.31\\

& 3D-UCaps \cite{nguyen20213ducaps} & 2021 &89.08 &90.67 & 94.54& 91.43& 0.47& 0.39& 0.14& 0.33\\   
& \textbf{Our 3DConvCaps}           & --   &\textbf{92.35} &\textbf{93.60} &\textbf{96.09} &\textbf{94.01} & \textbf{0.31}& \textbf{0.28}& \textbf{0.11}&  \textbf{0.23} \\ \bottomrule
    \end{tabular}
    \label{tab:testset}
\end{table*}

\begin{table}[th]
\centering 
\caption{Dice score comparison on Cardiac with 4-fold cross validation.}
\begin{tabular}{llccc}
\toprule
& Methods        & Year & DSC $\uparrow$ \\ \cline{2-4}
\multirow{3}{*}{\rotatebox{90}{\shortstack{3D CNNs\\-based MIS}}}
& 3D UNet\cite{cciccek20163d}  &  2016 &      84.30 \\ 
& 3D Vnet\cite{VNet} &     2018   &   84.20     \\
& 3D DR-UNet \cite{vesal2018dilated}   &  2018 &   87.40  \\ 
& HSSL \cite{zheng2021hierarchical} & 2021 & 87.65 \\
\midrule
\multirow{5}{*}{\rotatebox{90}{\shortstack{Capsule\\-based  MIS}}}
& SegCaps (2D)~\cite{lalonde2018capsules} & 2018  &  66.96  \\
& Multi-SegCaps (2D)~\cite{survarachakan2020capsule} & 2020  & 66.96\\
& 3D-UCaps \cite{nguyen20213ducaps} & 2021 & \underline{90.82} \\
%& SS-3DCapsNet \cite{tran2022ss} & 2022 & 89.77 \\
& \textbf{Our 3DConvCaps } & -- & \textbf{90.94 }\\ \bottomrule 
\end{tabular}
\label{table:cardiac}
\end{table}

% \begin{table}[]
% \centering
% \caption{Dice score comparison on Cardiac with 4-fold cross validation.}
% %\resizebox{\linewidth}{!}{
% \label{table:cardiac}
% \centering
% \begin{tabular}{ll|ll}
% \toprule
% \multicolumn{2}{l|}{3D CNNs-based networks} & \multicolumn{2}{l}{Capsule-based networks} \\ \cmidrule(l){1-4} 
% 3D UNet\cite{cciccek20163d}  &       84.30               & SegCaps (2D)~\cite{lalonde2018capsules}   &  66.96 \\ 
% 3D Vnet\cite{VNet} &          84.20            & Multi-SegCaps (2D)~\cite{survarachakan2020capsule}  & 66.96 \\
% 3D DR-UNet \cite{vesal2018dilated}   &     87.40    & 3D-UCaps \cite{nguyen20213ducaps} & 90.82 \\ 
% & & S-3DCapsNet \cite{tran2022ss} & 89.77 \\
% & &  \textbf{Our 3DConvCaps}  &       \textbf{90.94}          \\  
% \bottomrule
% \end{tabular}
% %}
% \end{table}

\begin{table}[th]
\centering
\caption{Dice score comparison on Hippocampus with 4-fold.}
% \label{table:hippocampus}
\begin{tabular}{l|ccc}
\toprule
Method & \multicolumn{3}{c} {Anterior/Posterior } \\ 
&  Recall$\uparrow$   & Precision$\uparrow$ & DSC $\uparrow$  \\ \midrule
%UNet (2.5D)\cite{}                       & 84.546   & 85.818    & 85.177 & 81.433    & 86.541    & 83.909 \\
Multi-SegCaps (2D)~\cite{survarachakan2020capsule}               & 80.76/84.46   & 65.65/60.49    & 72.42/70.49 \\
EM-SegCaps (2D)~\cite{survarachakan2020capsule}                  & 17.51/19.00   & 20.01/34.55    & 18.67/24.52 \\
%Ours 3DCapsNet                       & \textbf{81.70}   & \textbf{80.19}    & \textbf{80.99} & \textbf{80.20}    & \textbf{79.25}    & \textbf{79.48} \\ \bottomrule
3D-UCaps \cite{nguyen20213ducaps}                       & \textbf{94.88/93.59 } &\underline{77.48/74.03}  &\underline{85.07/82.49} \\ 
%SS-3DCapsNet \cite{tran2022ss} & 81.84/80.71 & 81.49/80.21 &  81.59/79.97\\
\textbf{Our 3DConvCaps}                       & \underline{88.66/85.36}  &  \textbf{87.9}/\textbf{87.94 }  & \textbf{88.06}/\textbf{86.47}  \\ \bottomrule
\end{tabular}
\label{table:hippocampus}
\end{table}

\begin{table*}[th]
\centering
\caption{Performance and network efficiency comparison on testing set of iSeg. }
\begin{tabular}{l|ccc|cc|cccc|cc}
\toprule
\multirow{2}{*}{Method} & \multicolumn{3}{c|}{Encoder} & \multicolumn{2}{c|}{Decoder} & \multicolumn{4}{c|}{Performance} & \multicolumn{2}{c}{Network Efficiency}\\ \cline{2-12}
& Conv. & Cap. & Con + Cap. & Conv. & Cap. & WM & GM & CSF & Aver. & Time Inf.(s) & Model size\\\hline
{3D-UNet  \mbox{\cite{cciccek20163d}}} & {\cmark} & {\xmark} & {\xmark} & {\cmark}& {\xmark} & {89.68}&{90.77} &{92.97} & {91.14}& {26}  &{5.8M} \\
3D-SegCaps \cite{nguyen20213ducaps} & \xmark & \cmark & \xmark & \xmark& \cmark & 71.01&73.60 &90.68 &78.43&  155 & 7.0M\\
3D-UCaps \cite{nguyen20213ducaps} &  \xmark & \cmark & \xmark & \cmark& \xmark &89.08 &90.67 & 94.54& 91.43 & 150 & 3.4M \\
\textbf{Our 3DConvCaps} &  \xmark & \xmark & \cmark & \cmark& \xmark & 92.35 &93.60 &96.09 &94.01 & 48 & 4.0M \\ \bottomrule
\end{tabular} 
\label{tb:ablation}
\end{table*}

\begin{figure*}[t]
    \centering
    \includegraphics[width=.85\textwidth]{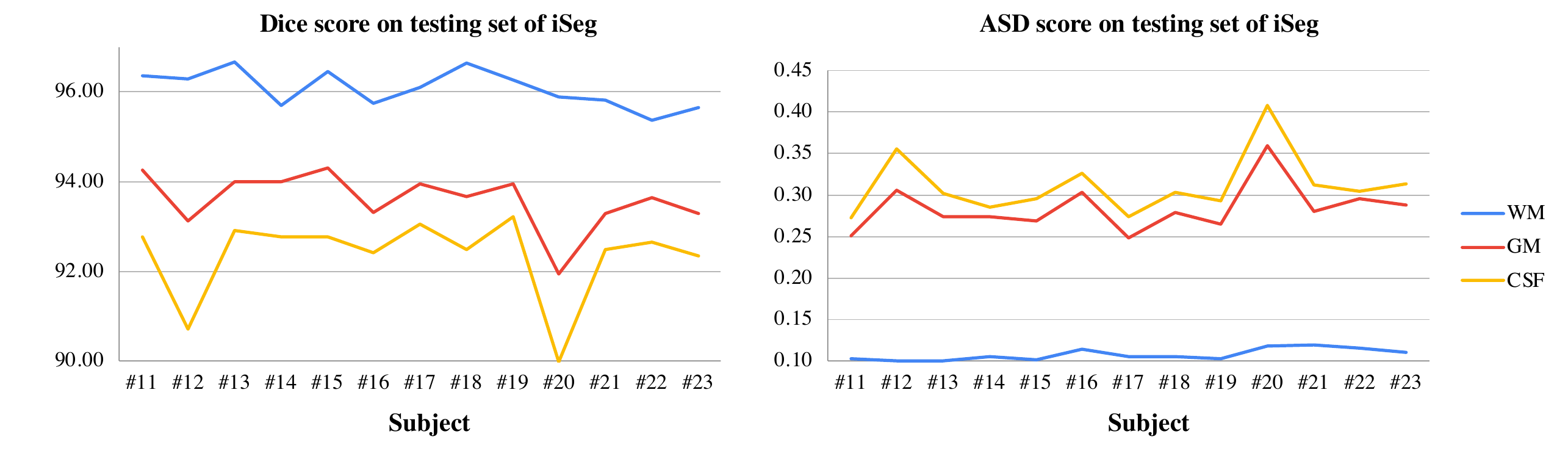}
    \label{fig:iseg-test}
    \caption{Performance of the proposed 3DConvCaps on the testing set with 13 subjects of iSeg-2017 datasets. Left: DSC score. Right: ASD metric.} 
\end{figure*}

\begin{figure*}[t]
    \centering
    \includegraphics[width=.85\textwidth]{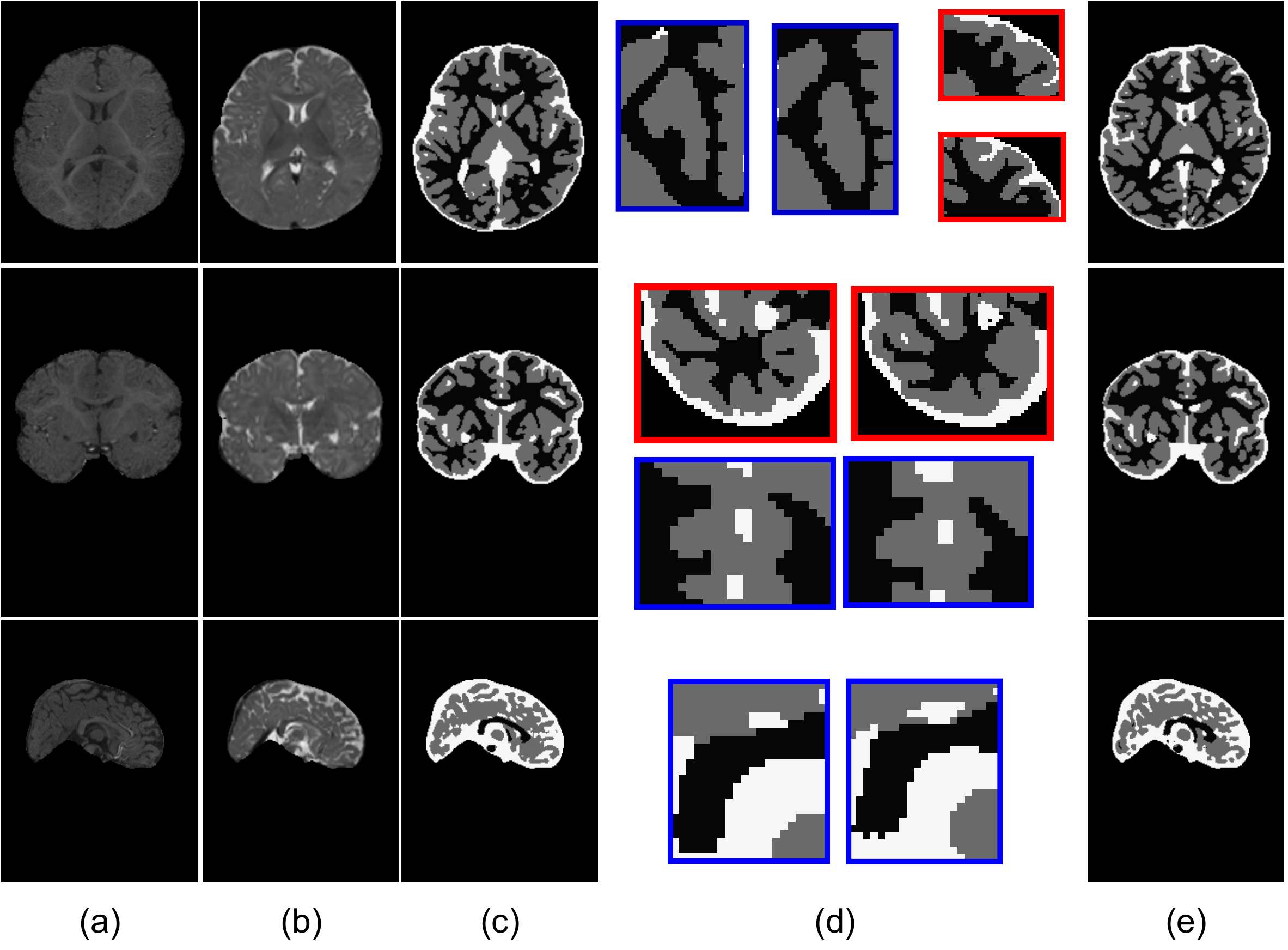}
    \label{fig:sub9}
    \caption{Visualization of validation subject i.e. subject \#9. (a) and (b) are T1- and T2-weighted brain MRI scans of subject \#9 in different views. (c): predicted segmentation results by our 3D ConvCaps; (d) Enlarged view of some random regions between (c) and (e). Blue boxes indicate some spots produced correctly by 3D ConvCaps. Some regions where 3D ConvCaps yielded incorrect segmentations are outlined in red ; (e): ground truth. Top-down: visualize in different planes: axial, coronal, and  sagittal.} 
\end{figure*}

% \begin{figure*}[t]
%     \centering
%     \includegraphics[width=.85\textwidth]{Figs/testset.pdf}
%     \label{fig:sub11-15}
%     \caption{Visualization of testing subjects i.e. subject \#11. Top-down: visualize in different planes: axial, coronal, and sagittal. For each subject, from left-right: T1-brain MRI, T2-brain MRI, predicted segmentation results conducted by our 3DConvCaps.} 
% \end{figure*}

\subsection{Loss function}
Our 3D ConvCaps network is supervised learnt by three loss functions as follows:
\begin{itemize}
    \item Margin loss: The loss is applied at the last layer in contracting path with downsampled ground truth segmentation. The margin loss is adopted from ~\cite{sabour2017dynamic} and it is defined between the predicted label $y$ and the ground truth label $y^*$ as follows:
    \begin{align}
    \mathcal{L}_{margin} =& y^* \times (\max(0, 0.9 - y))^2 + \\ & 0.5 \times (1 - y^*) \times (\max(0, y - 0.1))^2. \nonumber
    \end{align}
    \item Weighted cross entropy: The loss $\mathcal{L}_{CE}$ is applied at the last layer in expanding path to optimize the entire network.  
    \item Reconstruction: The loss $\mathcal{L}_{reconstruction}$ plays the role of network regularization and it is defined as the same as previous works \cite{lalonde2018capsules}. We use masked mean-squared error for this reconstruction loss. 
\end{itemize}
The total loss is the weighted sum of the three losses.
\begin{equation}
    \mathcal{L}_{downstream} = \mathcal{L}_{margin} + \mathcal{L}_{CE} + \mathcal{L}_{reconstruction}.
\end{equation}

\section{Experimental Results}
\subsection{Dataset}
\noindent \textbf{Infant brains} The iSeg17 dataset \cite{wang2019benchmark} consists of 10 subjects with ground-truth labels for training and 13 subjects without ground-truth labels for testing. Each subject includes T1 and T2 images with size of $144 \times 192 \times 256$, and image resolution of $1 \times 1 \times 1$  $\text{mm}^3$. In iSeg, there are three classes:  white matter (WM), gray matter (GM), and cerebrospinal fluid (CSF). 

\noindent \textbf{Heart:}. The Cardiac \cite{simpson2019large} is a mono-modal MRI dataset that contains 20 training images and 10 testing images. 

\noindent \textbf{Brain:} The Hippocampus \cite{simpson2019large} is a larger-scale mono-modal MRI dataset. It  consists of 260 training and 130 testing samples.
\subsection{Metrics:}
For quantitative assessment of the segmentation, the proposed model is evaluated on different metrics, e.g. Dice score (DSC), and average surface distance (ASD). 

The DSC measure is defined as:
\begin{equation}
    DSC = \frac{2|T \cap P|}{|T| + |P|}.
\end{equation}
where $T$ and $P$ are corresponding to groundtruth and predicted segmentation result. A higher value of DSC means better segmentation accuracy.

The ASD is utilized to measure the segmentation boundary distance and defined as:
\begin{equation}
\begin{split}
\small
    ASD(T, P) =&  \frac{1}{2}[\frac{\sum_{V_i\in S_P}{min_{V_j \in S_T}d(V_i, V_j)}}{\sum_{V_i\in S_P}{1}} \\ 
    & + \frac{\sum_{V_j\in S_T}{min_{V_i \in S_G}d(V_i, V_j)}}{\sum_{V_j\in S_T}{1}}].
\end{split}
\end{equation}
where $S_T$ and $S_P$ are the surface of the ground truth and predicted segmentation result. $d(V_i, V_j)$ is the Euclidean distance from a vertex $V_i$ and $V_j$. A smaller ASD is the better the result.

\subsection{Experiment Setup}
We perform our experiments and comparisons on iSeg-2017 \cite{wang2019benchmark}, Cardiac and Hippocampus \cite{simpson2019large}. Regarding iSeg-2017, we train the model on 10 annotated subjects and test on  13 unlabeled subjects (Subject \#11 - Subject \#23). Meanwhile, experiments on Cardiac and Hippocampus are conducted with 4-folds cross validation.

\subsection{Implementation details}

Based on 3D-UCaps \cite{nguyen20213ducaps} implementation, we implemented our methods with MONAI \cite{MonAI} for data management and Pytorch-lightning \cite{pytorch_light} for deep learning pipeline. The input volumes are normalized to [0, 1]. We used patch size set as $32 \times 32 \times 32$ for iSeg and Hippocampus whereas patch size set as $64 \times 64 \times 64$ on Cardiac. We set up our hyper-parameters as suggested in \cite{lalonde2018capsules}, where weight decay value for L2 regularization is 0.000002 and the learning rate is 0.0001. The learning rate is decayed by the ratio of 0.1 after 50,000 iterations without improvement on Dice score. Early stopping is set for 25,000 iterations. The experiments are conducted using an {Intel(R) Core(TM) i9-9920X CPU \@ 3.50GHz and GeForce RTX 2080 Ti.}

\subsection{Performance and Comparison}
\label{subsec:performance}
In this section, we compare our 3DConvCaps with both SOTA 3D CNNs-based MIS approaches and other existing Capsule-based MIS on three datasets. 

The quantitative comparison between our proposed 3DConvCaps with SOTA segmentation approaches on iSeg dataset \cite{wang2019benchmark} is given in Table~\ref{table:iseg}. Compared to both 3D CNNs-based MIS, Capsule-based MIS, our 3DConvCaps outperforms with a large margin on both DSC and ASD metrics compared to the second-best performance. For example, average DSC gains 1.01\% and average ASD reduces 0.03mm compared with the second-best \cite{zhuang2021aprnet} which is 3D CNNs-based approach. Compared to other Capsule-based approaches, our 3DConvCaps obtains the best score with a considerable gaps on both metrics i.e. average DSC gains 2.58\%  and average ASD reduces 0.11mm compared to the second-best 3D-UCaps\cite{nguyen20213ducaps}. Performance on DSC and ASD of the individual subject is provided in Fig. \ref{fig:iseg-test}. 

%The qualitative evaluation of 3DConvCaps on iSeg is shown in Fig. \ref{fig:sub9} and Fig. \ref{fig:sub11-15} corresponding to validation subject, i.e subject \#9 and some testing subjects i.e. subject \#11, \#13, \#15.

The qualitative evaluation of 3DConvCaps on iSeg is shown in Fig. \ref{fig:sub9} on validation subject, i.e subject \#9.

On Cardiac dataset, Dice score performance comparison is reported in Table \ref{table:cardiac}, which shows that our method does not only outperforms other Capsule-based MIS approaches but also 3D CNNs-based MIS ones with a large margin. 

On Hippocampus dataset, we report both Dice score, Precision and Recall metrics on both Anterior and Posterior as shown in Table \ref{table:hippocampus}. While our 3DConvCaps obtains the second-best on Recall, it gains the best performance on Precision and Dice score with a big difference. With higher precision, it means that our approach returns more relevant results than irrelevant ones. For example, average DSC gains 3.73\% and average Precision increases 12.2\% compared to the second-best \cite{nguyen20213ducaps}.

We further examine the balance between performance and network efficiency i.e. time inference and model size of our 3DConvCaps compared to other {CNN-based and} Capsule-based MIS approaches as shown in Table \ref{tb:ablation}.
{In this table, we implement 3DUNet by replacing all capsule layers in our 3DConvCaps with convolutional layers as a baseline. This baseline also shares the same number of layers
and blocks as other capsule-based networks in Table \mbox{\ref{tb:ablation}}.
As shown in Table \mbox{\ref{tb:ablation}}, 3D-UCaps \mbox{\cite{nguyen20213ducaps}}, a SOTA 3D Capsule-based approach in medical image segmentation, obtains competitive results with 3D-UNet (baseline) with less model size. However, time inference consumption is a big challenge in 3D-UCaps \mbox{\cite{nguyen20213ducaps}} because of dynamic routing. Compared to 3D-UNet (baseline), our 3DConvCap yields a substantial margin on performance while keeping model size smaller and time inference plausible. In comparison with the SOTA 3D Capsule-based approach, 3D-UCaps \mbox{\cite{nguyen20213ducaps}}, our network not only outperforms on all three classes of WM, GM, CSF but also requires less time inference with competitive model size.}

% Designed by capsule layers only at both contracting and expanding paths, 3D-SegCaps~\cite{nguyen20213ducaps} is the most expensive computation and lowest performance. {Meanwhile, our method shows efficiency by leveraging both CNN layers and capsule layers for short-range and long-range information, respectively. Indeed, as in Table. \mbox{\ref{tb:ablation}}, 3DConvCaps outperforms other methods by performance. In addition, our method has acceptable time inference, compared to the fastest one 3D-UNet (i.e. 48s v.s. 26s), and has a comparable model size, in comparison with the smallest one 3D-UCaps (i.e. 4.0M v.s. 3.4M).} 
% \sout{While our 3DConvCaps' model size is compatible with 3D-UCaps, our computation is much more effective i.e. 48s v.s. 150s at inference time.}

\section{Conclusion}
In this paper, we have proposed an efficient network architecture that inherits the merits of both capsule layers to model the part-whole relationship at higher-level features and convolutional layers to learn short-range information at lower-level features. Compared to other SOTA Capsule-based and 3D CNNs-based MIS approaches, our 3DConvCaps is more effective and robust in terms of both performance and network efficiency. 

Future work includes analyzing the robustness on various transformations (e.g rotation) invariance and motion artifact. We also further explore other recent techniques such as self-supervised learning and supervised contrastive learning for better feature learning and representation.

\section*{Acknowledgments}
This material is based upon work supported in part by the US National Science Foundation, under Award No. OIA-1946391, NSF 1920920, NSF Data Science, Data Analytics that are Robust and Trusted (DART) and NSF WVAR-CRESH Grant.
The authors would like to thank the reviewers for their valuable comments.

%\section*{Disclaimer}
%Any opinions, findings, and conclusions or recommendations expressed in this material are those of the author(s) and do not necessarily reflect the views of the National Science Foundation.

\bibliographystyle{IEEEtran}
\bibliography{refs}

% \newpage
% \input{author_response}
\end{document}